\documentclass[10pt,twocolumn,letterpaper]{article}

\usepackage[pagenumbers]{wacv} % To force page numbers, e.g. for an arXiv version

\usepackage{graphicx}
\usepackage{amsmath}
\usepackage{amssymb}
\usepackage{booktabs}
\usepackage{textcomp}
\usepackage{stfloats}
\usepackage{url}
\usepackage{verbatim}
\usepackage{graphicx}
\usepackage{cite}

\usepackage{booktabs}
\usepackage{bbm}
\usepackage{mathtools}
\DeclareMathOperator{\Tr}{Tr}
\usepackage{footnote}

\usepackage{amsthm}
\usepackage{array}
\usepackage{tabularx}
\usepackage{threeparttable}
\DeclareMathOperator*{\dprime}{\prime \prime}
\usepackage{epstopdf} %converting to PDF
\makeatother

\usepackage{bbm}

\usepackage{algorithm}
\usepackage{listings}

\usepackage[pagebackref,breaklinks,colorlinks]{hyperref}

\usepackage[capitalize]{cleveref}
\crefname{section}{Sec.}{Secs.}
\Crefname{section}{Section}{Sections}
\Crefname{table}{Table}{Tables}
\crefname{table}{Tab.}{Tabs.}

\def\wacvPaperID{1110} % *** Enter the WACV Paper ID here
\def\confName{WACV}
\def\confYear{2024}

\begin{document}

\title{Universal Semi-supervised Model Adaptation via Collaborative \\Consistency Training}

% \and
\author{
Zizheng Yan\textsuperscript{\rm 1,2}\thanks{Equal contributions.} \quad Yushuang Wu\textsuperscript{\rm 1,2}\footnotemark[1] \quad Yipeng Qin\textsuperscript{\rm 3} 
\quad Xiaoguang Han\textsuperscript{\rm 1,2}  \\
\quad Shuguang Cui\textsuperscript{\rm 1,2} \quad Guanbin Li\textsuperscript{\rm 4,5}\thanks{Corresponding Author.}
\vspace{5pt}\\
\textsuperscript{\rm 1}{FNii, CUHKSZ} \qquad \textsuperscript{\rm 2}{SSE, CUHKSZ} \qquad
\textsuperscript{\rm 3}{Cardiff University} \quad
\textsuperscript{\rm 4}{Sun Yat-sen University} \\
\textsuperscript{\rm 5}{Research Institute, Sun Yat-sen University, Shenzhen}
% {\tt\small \{zizhengyan, yushuangwu\}@link.cuhk.edu.cn} \quad
% {\tt\small qiny16@cardiff.ac.uk}  \\
% {\tt\small \{shuguangcui, hanxiaoguang\}@cuhk.edu.cn} \quad
% {\tt\small liguanbin@mail.sysu.edu.cn}
}

\maketitle
%%%%%%%%% ABSTRACT
\begin{abstract}
   In this paper, we introduce a realistic and challenging domain adaptation problem called Universal Semi-supervised Model Adaptation (USMA), which i) requires only a pre-trained source model, ii) allows the source and target domain to have different label sets, i.e., they share a common label set and hold their own private label set, and iii) requires only a few labeled samples in each class of the target domain. To address USMA, we propose a collaborative consistency training framework that regularizes the prediction consistency between two models, i.e., a pre-trained source model and its variant pre-trained with target data only, and combines their complementary strengths to learn a more powerful model. The rationale of our framework stems from the observation that the source model performs better on common categories than the target-only model, while on target-private categories, the target-only model performs better. We also propose a two-perspective, i.e., sample-wise and class-wise, consistency regularization to improve the training. Experimental results demonstrate the effectiveness of our method on several benchmark datasets.
\end{abstract}

%%%%%%%%% BODY TEXT
\section{Introduction}

\begin{figure}[t]
    \centering
    \vspace{2mm}
    \includegraphics[width=0.99\linewidth]{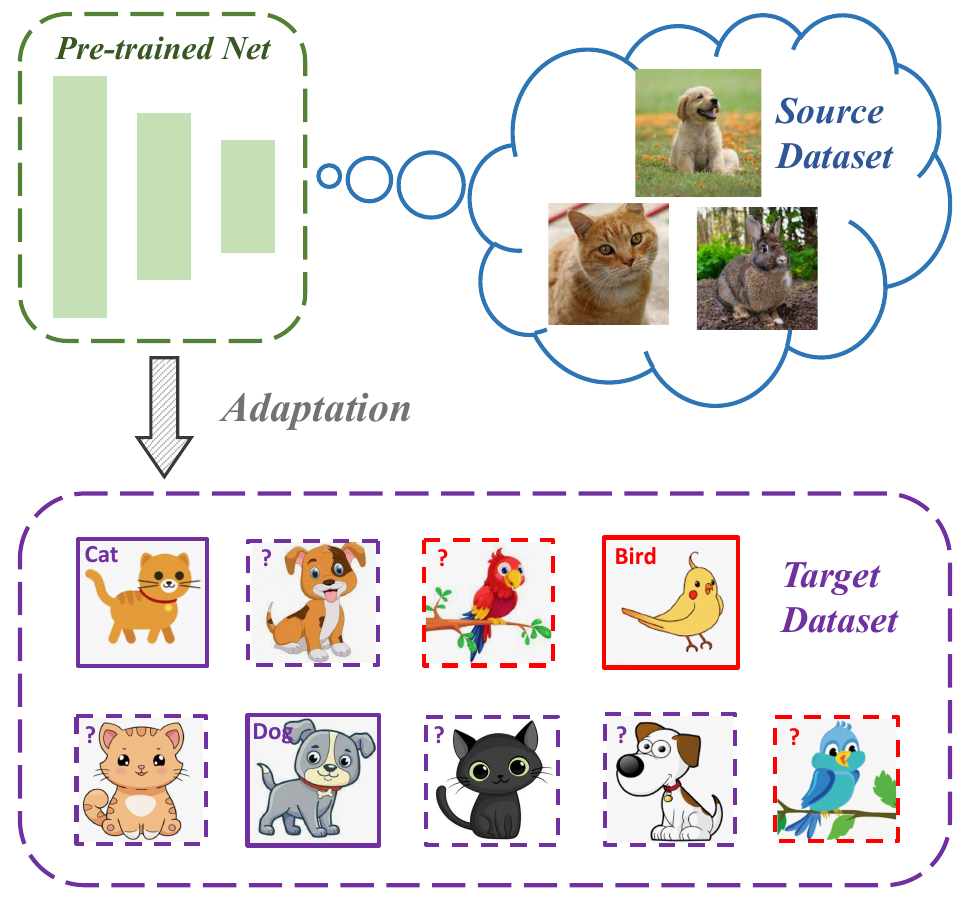}\\
    \caption{
    Illustration of Universal Semi-supervised Model Adaptation (USMA). In this example, we begin with a source model that has been pre-trained on a real-world domain dataset containing categories "Cat", "Dog", and "Rabbit". The target domain consists of a cartoon dataset with different categories,  "Cat", "Dog", and "Bird", each containing both labeled~(Solid boxes) and unlabeled samples~(Dashed boxes). 
    }
    \vspace{-4mm}
    \label{fig:teaser}
\end{figure}

Deep neural networks have achieved remarkable progress in various tasks, such as image recognition~\cite{resnet}, machine translation~\cite{translation}, biomedical imaging~\cite{unet}, etc. 
However, training a good neural network model remains challenging as it requires huge amounts of labeled data that are expensive to annotate.
To this end, unsupervised domain adaptation (UDA) methods~\cite{grl,tzeng2017adversarial,yan2021pixel} were proposed to train neural network models without annotated data by transferring knowledge learned from a label-rich source domain to the unlabeled target domain. 
However, due to the gap between the source and target domains, it is still challenging for UDA methods to achieve good performance on the target domain.
To bridge such a domain gap, Semi-supervised Domain Adaptation (SSDA)~\cite{saito2019semi,li2021cross} incorporates a few labeled target samples into the training and thus significantly outperforms UDA, showing great potential for applications.

Although promising, existing SSDA methods usually assume that the source data is available during training, which is impractical in many real-world scenarios where restrictions apply, \eg,~data privacy and limited storage~\cite{shot}.
To meet such new demands, a new research topic, namely the model adaptation~\cite{shot,yang2020bait,liu2021source,ma2021semi,wang2021learning}, has recently been proposed with the aim of transferring knowledge from a pre-trained source model rather than the source data. 
To simplify the problem, most of these works assume that the source and target domain share the same label set. 

However, the above simplification significantly limits the practical application of those methods due to the prevalence  of category gaps between the pre-trained source models and the target data.
For example, when adapting a product retrieval model pre-trained on a web dataset to real-world ones collected from shelves of different supermarkets, there are likely different subsets of common categories between the pre-trained source model and each real-world dataset.
This issue is addressed by universal domain adaptation~\cite{you2019universal}. 

\begin{figure}[t]
% \captionsetup{labelformat=empty}
\centering
\includegraphics[width=.99\linewidth]{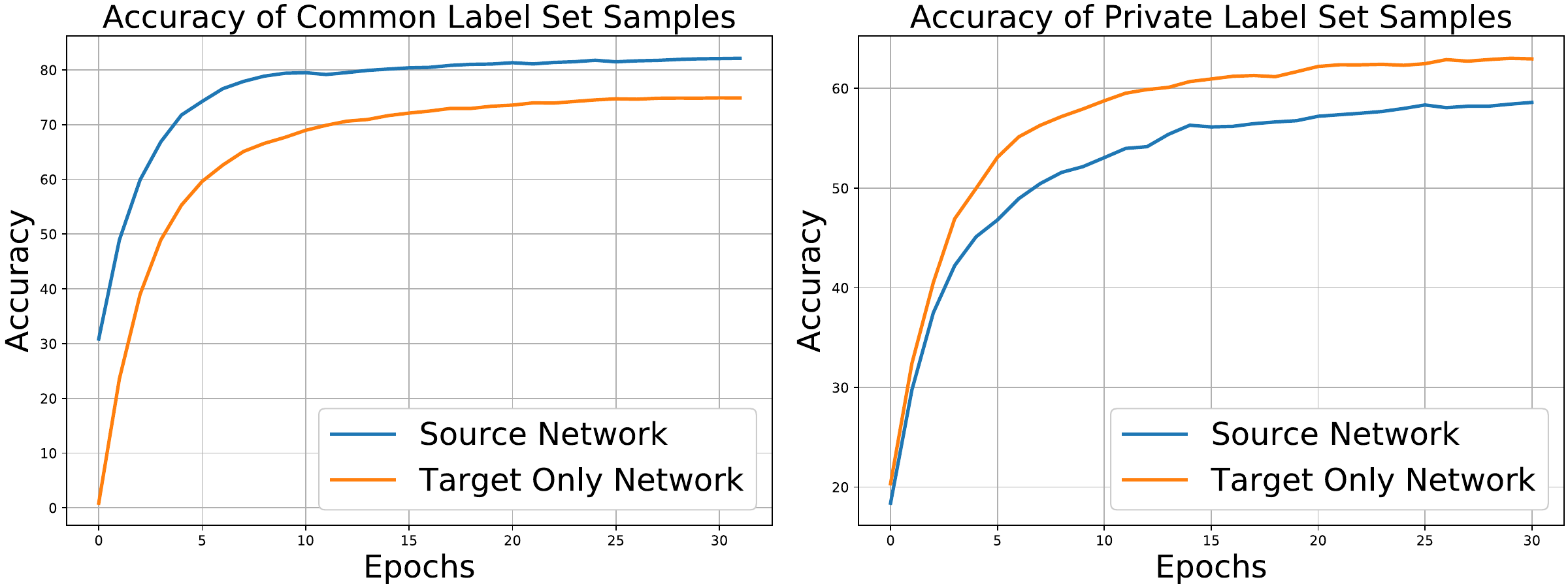}
\caption{Left: The classification accuracy of samples from common label set. Right: The classification accuracy of the samples from private label set. The blue curve shows the accuracy of the network fine-tuning from the source pre-trained network while the orange curve shows the accuracy of the network fine-tuning from a network pre-trained with only target data through self-supervised learning.}
\label{fig:common_private_acc}
\vspace{-4mm}
\end{figure}

In this paper, we propose a new problem, called \textit{Universal Semi-supervised Model Adaptation} (USMA), which covers all aforementioned problems. USMA presents a more realistic and, consequently, more challenging scenario. An Illustration of USMA is shown in Fig.~\ref{fig:teaser}. 
A naive solution for USMA is to apply a semi-supervised learning method to fine-tune the pre-trained source model directly with the target data. 
However, we observed that such a naive solution is ineffective as the source model's extensive knowledge in the source domain impedes its learning in the target domain: 
it works better for common categories shared by both the source and target domains, than the target-private categories (\ie,~categories in the target but not in the source domain).
This is justified by the opposite performance of a reference model, \ie,~a model with the same architecture of the source model but pre-trained using only the target data in a self-supervised way (Fig.~\ref{fig:common_private_acc}).
Despite such a disappointment, the complementary strengths of the two models caught our attention: can we integrate the appeal of the two to train a more powerful model that performs well on both common and target-private categories?

To answer the above question, we propose a collaborative consistency training (CCT) framework which extends the vanilla consistency regularization applicable to a single model~\cite{xie2019unsupervised,sohn2020fixmatch} to our double-model case.
Specifically, we propose to add additional regularization across the two models, thereby allowing each model to exploit the strengths of the other model to resolve its shortcomings.
To take full advantage of the proposed CCT framework, we further propose to incorporate consistency regularization from two perspectives, \ie,~sample-wise and class-wise.
Sample-wise, we first augment each training sample into two views and employ pseudo labeling to enforce consistency of high-confidence predictions between not only the two views of a single model but also those across two different models. 
Class-wise, we propose a single loss function that incorporates both the \textit{class consistency prior} (\ie,~same prediction across views) and the \textit{class sparsity prior} (\ie,~sparsity of prediction vectors) into training.

Interestingly, we observed that the performance of the two models converge to a similar point after training. Therefore, without loss of generality, we choose the adapted source model as the final model for reference.
Our contributions include:
\begin{itemize}
    \vspace{-2.5mm}
    \item We define Universal Semi-supervised Model Adaptation (USMA), a more realistic and challenging domain adaptation problem to be solved.
    \vspace{-2.5mm}
    \item We propose a collaborative consistency training~(CCT) framework that leverages the complementary strengths of a source-pretrained and target-pretrained model to produce a more powerful one.
    \vspace{-2.5mm}
    \item We propose a two-perspective approach (sample-wise and class-wise) for CCT. 
    \vspace{-2.5mm}
    \item Extensive experimental results demonstrate the effectiveness of the proposed methods.
\end{itemize}

\section{Related Work}

\noindent
\textbf{Semi-supervised Domain Adaptation.} Semi-supervised Domain Adaptation (SSDA) assumes few labeled samples per category in the target domain~\cite{saito2019semi,yang2020mico,qin2021contradictory,li2021ecacl,kim2020attract,li2021cross,jiang2020bidirectional,li2020online,liang2021adc,li2021invrisk,yan2022multi,berthelot2022adamatch,wu2023scoda}, which yields cost-effective applications that require little labeling efforts.
Most of those works address the SSDA problem from two perspectives: (i) domain alignment and (ii) self-training on target domain, \ie,~entropy minimization, pseudo labeling. Saito~\etal~\cite{saito2019semi} first introduce the SSDA problem and propose to minimize entropy for feature extractor and align the source and target domain by maximizing entropy for the classifier. UODA~\cite{qin2021contradictory} proposes to minimize the target entropy so that the target feature can be compact, and maximize the source entropy so that the source feature can be scattered, making alignment easier. CDAC~\cite{li2021cross} proposes an adaptive clustering method to perform domain alignment and uses pseudo labeling for self-training. ECACL~\cite{li2021ecacl} proposes to reduce the domain gap by matching source and target prototypes. AdaMatch~\cite{berthelot2022adamatch} proposes to match the class distribution between source and target.
Compared with the vanilla SSDA, our proposed USMA is more challenging as it has no access to source data, which is infeasible to perform domain alignment, and also needs to adapt models to a target domain that has a different label set than the source domain.

\noindent
\textbf{Model Adaptation.}
Model adaptation (MA) was proposed to implement domain adaptation without access to the source data, thereby addressing the dilemma of data sharing versus data privacy~\cite{shot} in traditional domain adaptation.
Many works have been proposed to address the problem of MA~\cite{shot,li2020model,yang2020bait,xia2021adaptive,yeh2021sofa,ma2021semi,wang2021learning,kurmi2021domain,kundu2020universal,yang2021generalized,chi2021tohan,shotplus,ding2022source,kundu2022concurrent,qu2022bmd,roy2022uncertainty}, and the methods can be roughly categorized into two streams: generative~\cite{li2020model,yeh2021sofa} and discriminative~\cite{shot,yang2020bait,xia2021adaptive}. Generative methods usually model the generation of labeled images or features. 3C-GAN~\cite{li2020model} proposes to synthesize the target-style labeled training images via conditional GAN. SoFA~\cite{yeh2021sofa} proposes to generate reliable latent features for domain alignment. On the other hand, discriminative methods usually require to fix the source classifier and fine-tune the backbone. SHOT~\cite{shot,shotplus} proposes to minimize the target entropy while maximizing the mutual information. AANet~\cite{xia2021adaptive} proposes to incorporate a trainable classifier with the fixed classifier to jointly perform adaptation. 
Though achieving good performance in unsupervised or semi-supervised settings, these MA methods work under the assumption that the source and target domains share the same label set, which is not the case in USMA.

\noindent
\textbf{Universal Domain Adaptation.} 
Universal DA~\cite{you2019universal,fu2020learning,saito2020universal,saito2021ovanet,Ma2021acuda,li2021domain,deng2021universal,yu2021divergence,lifshitz2020sample,Chen_2022_universal} resolves the class mismatch issue by combining Open-set DA~\cite{saito2018open} and Partial DA~\cite{cao2018partial}. Open-set DA excludes some target domain categories (a.k.a. target-private categories) not in the source domain, while Partial DA excludes some source domain categories (a.k.a. source-private categories) not in the target domain. 
Most works focus on designing the criterion to reject target-private samples. UAN~\cite{you2019universal} proposes to use entropy as the criterion, \eg,~reject high entropy samples and exclude them from domain alignment. Fu~\etal~\cite{fu2020learning} propose the joint use of entropy, confidence and classifier consistency as the criterion. DANCE~\cite{saito2020universal} also uses entropy as the criterion and proposes an entropy separation loss to reject the target-private samples. OVANet~\cite{saito2021ovanet} trains one-vs-all classifiers to learn the distance between the source positive and nearest negative classes and use such distance as the criterion.
Instead of rejecting open-set samples as in universal DA, USMA aims to classify them correctly in a semi-supervised way given few labeled samples in the target-private categories, which is more challenging.

\section{Methodology}

\begin{figure*}[t]
    \centering
    % \vspace{2mm}
    \includegraphics[width=1\linewidth]{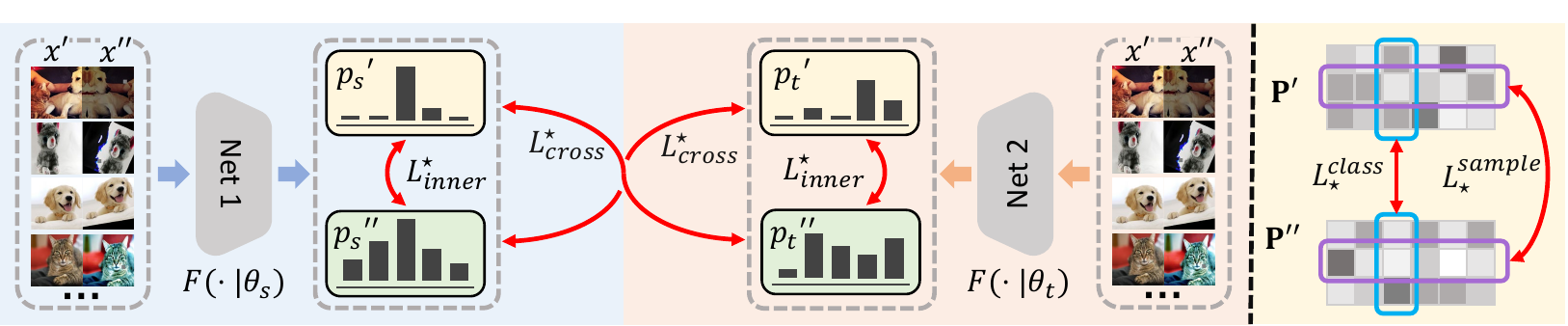}\\
    \caption{Illustration of the proposed collaborative consistency training (CCT) framework. Left: the overall pipeline of CCT. Right: The detailed consistency loss of CCT. The backbone of $F(\cdot|\theta_s)$ is initialized with the source-pretrained model, while the backbone of $F(\cdot|\theta_t)$ is initialized with the self-supervised pre-trained model using target data. $\mathbf{P}^{\prime}, \mathbf{P}^{\dprime} \in \mathbbm{R}^{N \times C}$ are the batch predictions of two views, where $N$ is the batch size and $C$ is the number of classes.
    }
    \label{fig:overview}
    \vspace{-2mm}
\end{figure*}

In this section, we first present the definition of USMA, and then detail the proposed collaborative consistency training framework. Finally, we propose two novel loss functions that regularize collaborative consistency in a  sample-wise and class-wise manner respectively. 
Our framework is illustrated in Fig.~\ref{fig:overview}.

\subsection{Problem Definition}
\label{sec:problem_definition}

Universal Semi-supervised Model Adaptation (USMA) aims to learn a model $F$ that achieves high prediction accuracy on target domain data using only $F_s$ and $\mathcal{D}_t$, where 
\begin{itemize}
    \vspace{-2mm}
     \item $F_s$ denotes a source model pre-trained with the source data $\mathcal{D}_s = \{(\mathbf{x}_i^s, \mathbf{y}_i^s)\}, \mathbf{y}_i^s \in \mathcal{C}_s$ sampled from a different distribution than the target domain,
     \vspace{-2mm}
    \item $\mathcal{D}_t = \mathcal{D}_t^l \cup \mathcal{D}_t^u = \{(\mathbf{x}_i^{tl}, \mathbf{y}_i^{tl})\}_{i=1}^{n_{tl}} \cup \{(\mathbf{x}_j^{tu})\}_{j=1}^{n_{tu}}, \mathbf{y}_i^{tl} \in \mathcal{C}_t$ denotes a target dataset consisting of $n_{tl}$ labeled and $n_{tu}$ unlabeled samples ($n_{tl} \ll n_{tu}$),
    \vspace{-2mm}
    \item Label sets $\mathcal{C}_s \neq \mathcal{C}_t$.
    \vspace{-2mm}
\end{itemize}

To facilitate discussion, we denote $\mathcal{C} = \mathcal{C}_s \cap \mathcal{C}_t$ as the common label set, and have $\mathcal{\Bar{C}}_s = \mathcal{C}_s \backslash \mathcal{C}$ and $\mathcal{\Bar{C}}_t = \mathcal{C}_t \backslash \mathcal{C}$ that denote the label sets of private categories of source and target domains respectively. Following~\cite{you2019universal}, we define the commonness of label sets $\xi$ as $ \frac{|\mathcal{C}_s \cap \mathcal{C}_t|}{|\mathcal{C}_s \cup \mathcal{C}_t|}$.

\subsection{Collaborative Consistency Training}

As Fig.~\ref{fig:overview} shows, the key insight of our framework is that the final model $F(\cdot|\mathbf{\theta})$ can be derived from the collaborative training of $F(\cdot|\mathbf{\theta}_s)$ and $F(\cdot|\mathbf{\theta}_t)$ that enforces consistent predictions in-between, where $F(\cdot|\mathbf{\theta}_s) = F_s$ is the model pre-trained with source data~({\it a.k.a.}~source model) and $F(\cdot|\mathbf{\theta}_t)$ is the target model pre-trained with $\mathcal{D}_t$ in a self-supervised learning manner~({\it a.k.a.}~target only model).
Both $F(\cdot|\mathbf{\theta}_s)$ and $F(\cdot|\mathbf{\theta}_t)$ share the same network architecture but have different parameters $\theta_s$ and $\theta_t$. Specifically, $F(\cdot|\mathbf{\theta})$ consists of a backbone and a classifier. The backbones of $F(\cdot|\mathbf{\theta}_s)$ and $F(\cdot|\mathbf{\theta}_t)$ are pretrained with different data, while the classifiers are both randomly initialized. The output dimension of the classifiers of $F(\cdot|\mathbf{\theta}_s)$ and $F(\cdot|\mathbf{\theta}_t)$ is $|\mathcal{C}_t|$.
The rationale of our approach stems from the observation (Fig.~\ref{fig:common_private_acc}) that: i) for samples with common labels $\mathcal{C}$, despite the domain gaps, $F(\cdot|\mathbf{\theta}_s)$ outperforms $F(\cdot|\mathbf{\theta}_t)$ in prediction accuracy from its rich prior knowledge in $\mathcal{C}$; ii) while for samples with target private labels $\mathcal{\Bar{C}}_t$, $F(\cdot|\mathbf{\theta}_t)$ achieves a higher prediction accuracy than $F(\cdot|\mathbf{\theta}_s)$.
Such an observation implies that $F(\cdot|\theta_s)$ and $F(\cdot|\theta_t)$ learn complementary information of $\mathcal{C}$ and $\mathcal{\Bar{C}}_t$ ($\mathcal{C}_t = \mathcal{C} \cup \mathcal{\Bar{C}}_t$ ) in the target domain respectively.

To harness this, we introduce a Collaborative Consistency Training (CCT) framework. This involves creating dual views of each sample in the unlabeled target data $\mathcal{D}_t^u$ through augmentations. Unlike traditional consistency regularization, which enforces internal consistency within a single model, we introduce \textit{cross consistency}. This aligns predictions between the two models $F(\cdot|\mathbf{\theta}_s)$ and $F(\cdot|\mathbf{\theta}_t)$, enabling them to learn more discriminative features for both common and private classes.
We further enrich CCT by implementing both inner and cross consistency from sample-wise and class-wise perspectives. Subsequent sections detail these regularization techniques.

\subsection{Sample-wise Consistency Regularization}

For each unlabeled target sample $\mathbf{x}_i \in \mathcal{D}_t^u $, we apply two different augmentations to it and get its two views $\mathbf{x}_i^{\prime}$ and  $\mathbf{x}_i^{\dprime}$, respectively.

\noindent \textbf{Sample-wise Inner Consistency Regularization.}
~~We follow the vanilla consistency regularization in SSL~\cite{xie2019unsupervised,sohn2020fixmatch} and use the pseudo labels generated by one sample view $\mathbf{x}_i^{\prime}$ to supervise the training of the other sample view $\mathbf{x}_i^{\dprime}$.
For each sample, the loss function can be formulated as:
\begin{equation}
\vspace{-2mm}
 \begin{aligned}
\mathcal{L}_{sample}^{inner} = &-\sum_{i=1}^{C}\mathbbm{1}(F(\mathbf{x}^{\prime}|\theta_s)_i \ge \tau) \log (F(\mathbf{x}^{\dprime}|\theta_s)_i)\\
&-\sum_{i=1}^{C}\mathbbm{1}(F(\mathbf{x}^{\prime}|\theta_t)_i \ge \tau) \log (F(\mathbf{x}^{\dprime}|\theta_t)_i),
\label{eq:pl}
\end{aligned}   
\vspace{-2mm}
\end{equation}
where $C = |\mathcal{C}_t|$ is the number of classes in the target domain, $\tau$ is the threshold. 
As indicated in~\cite{xie2019unsupervised}, $\mathcal{L}_{sample}^{inner}$ regularizes both networks to be invariant to input noises (\ie, different views) and makes them more robust.

\noindent \textbf{Sample-wise Cross Consistency Regularization.}
While in USMA, the aforementioned inner consistency regularization is not sufficient as it suffers from the poor performance of the source model $F(\cdot|\theta_s)$ on target private categories $\mathcal{\Bar{C}}_t$.
Our sample-wise cross consistency regularization addresses this issue by collaboratively training $F(\cdot|\theta_s)$ and the target model $F(\cdot|\theta_t)$ together:
\begin{equation}
\vspace{-2mm}
\begin{aligned}
\mathcal{L}_{sample}^{cross} = & -\sum_{i=1}^{C}\mathbbm{1}(F(\mathbf{x}^{\prime}|\theta_s)_i \ge \tau) \log (F(\mathbf{x}^{\dprime}|\theta_t)_i)  \\ 
 & - \sum_{i=1}^{C}\mathbbm{1}(F(\mathbf{x}^{\prime}|\theta_t)_i \ge \tau) \log (F(\mathbf{x}^{\dprime}|\theta_s)_i), 
\label{eq:sample_cross}
\end{aligned}
\vspace{-2mm}
\end{equation}
where the highly-confident predictions of $F(\mathbf{x}^{\prime}|\theta_t)$ 
are used as pseudo labels of $F(\mathbf{x}^{\dprime}|\theta_s)$ and the highly-confident predictions of $F(\mathbf{x}^{\prime}|\theta_s)$ (mostly samples of $\mathcal{C}$) are used as pseudo labels of $F(\mathbf{x}^{\dprime}|\theta_t)$.

\noindent \textbf{Sample-wise Consistency Loss.}
Combining the inner and cross consistency loss, we have our final sample-wise consistency loss as:
\begin{equation}
\begin{aligned}
\mathcal{L}_{sample} &= \frac{1}{2}(\mathcal{L}_{sample}^{inner} + \mathcal{L}_{sample}^{cross}).
\label{eq:sample_v1}
\end{aligned}
\end{equation}

\noindent \textbf{Remark. }
The above objective indicates that each sample receives supervision from \textit{two pseudo labels}.
Since the two networks have learned complementary information, we argue that it is highly likely that at least one of them is correct.
Furthermore, as indicated by \cite{arpit2017closer} that DNNs tend to fit clean labels before noisy ones, our networks will prioritize the fitting of such correct pseudo labels.
This helps our networks learn better representations and provides more reliable pseudo labels for subsequent training steps, thus forming a virtuous circle of training.

\subsection{Class-wise Consistency Regularization}
\label{subsec:classwise}
Being orthogonal to our sample-wise consistency regularization, our class-wise consistency regularization facilitates the proposed CCT through the incorporation of {\it class consistency prior} and {\it class sparsity prior} that takes the relation among samples in a batch into consideration.
Similar to the above section, we denote $\mathbf{x}_i^{\prime}$ and $\mathbf{x}_i^{\dprime}$ as the two views of an unlabeled target sample $\mathbf{x}_i \in \mathcal{D}_t^u$ and have: 1) \textbf{Class Consistency Prior.} The predictions of $\mathbf{x}_i^{\prime}$ and $\mathbf{x}_i^{\dprime}$ should be the same over all classes. 2) \textbf{Class Sparsity Prior.} The predictions of $\mathbf{x}_i^{\prime}$ and $\mathbf{x}_i^{\dprime}$ should be unit vectors with a single $1.0$ for one class and $0$ for all the other classes.

\noindent \textbf{Class-wise Inner Consistency Regularization.}~~~~Suppose $\mathbf{P}^{\prime}, \mathbf{P}^{\dprime} \in \mathbb{R}^{n \times C}$ are the batch predictions of $F(\mathbf{x}^{\prime}|\theta_s)$ and $F(\mathbf{x}^{\dprime}|\theta_s)$ respectively, where $n$ is the batch size, $C$ is the number of classes, and $\mathbf{P}_{(j,k)}$ represents the softmax confidence of classifying the $j$th sample into the $k$th class, we 
compute the correlation matrix of $\mathbf{P}^{\prime}$ and $\mathbf{P}^{\dprime}$ as: 
\begin{align}
\vspace{-2mm}
\mathbf{\mathbf{R}}= {\mathbf{P}^{\prime}}^\top\mathbf{P}^{\dprime}.
\vspace{-2mm}
\end{align}
It can be observed that $\mathbf{\mathbf{R}}_{(j, k)}$ represents the inner product similarity between the $j$-th and $k$-th column of $\mathbf{P}^{\prime}$ and $\mathbf{P}^{\dprime}$, respectively. 
Such a cross-correlation matrix $\mathbf{R}$ allows us to incorporate i) the \textbf{class consistency prior} by maximizing the similarity between the same classes (diagonal values) and ii) the \textbf{class sparsity prior} by minimizing the similarity between different classes (off-diagonal values). 
\footnote{This works as when two columns are similar, there will be two non-zero values (two classes) for one sample, which violates the sparsity prior.}

Since $\mathbf{R}$ is asymmetric and the row summations are different, we convert it to be symmetric and normalize its rows and columns:
\begin{align}
\mathbf{\mathbf{\hat{R}}} = \phi\big((\mathbf{\mathbf{R}} + \mathbf{\mathbf{R}}^\top) / 2\big),
\label{eq:nomarlize}
\end{align}
where $\phi(\cdot)$ represents a row normalization where each element is divided by the row sum. With $\mathbf{\hat{R}}$, we can formulate our class-wise inner consistency loss as:
\begin{align}
\mathcal{L}_{class}^{inner} = -\frac{1}{2C}\big(\Tr({\mathbf{\hat{R}}_s}) + \Tr({\mathbf{\hat{R}}_t})\big),
\label{eq:class_inner}
\end{align}
where $\Tr(\cdot)$ is the trace of a matrix, $\mathbf{\hat{R}}_s$ and $\mathbf{\hat{R}}_t$ are the computed matrices from $F(\cdot|\theta_s)$ and $F(\cdot|\theta_t)$, respectively. 
With the normalization in Eq.~\ref{eq:nomarlize} (\ie, the sum of values in each row or column equals to 1), minimizing the above loss function not only maximizes the diagonal values but also minimizes the off-diagonal values, which incorporates the proposed two priors simultaneously. 

\vspace{1mm}\noindent \textbf{Class-wise Cross Consistency Regularization.} We extend the above inner consistency regularization to a cross consistency one as follows. 
Let $\mathbf{P}_1^{\prime}, \mathbf{P}_2^{\dprime} \in \mathbb{R}^{n \times C}$ be the batch predictions of $F(\mathbf{x}^{\prime}|\theta_s)$ and $F(\mathbf{x}^{\dprime}|\theta_t)$ respectively, we compute the cross-correlation matrices as: 
\begin{align}
\mathbf{\mathbf{R}}_1 = {\mathbf{P}_1^{\prime}}^\top\mathbf{P}_2^{\dprime}, \quad
\mathbf{\mathbf{R}}_2 = {\mathbf{P}_2^{\prime}}^\top\mathbf{P}_1^{\dprime}.
\end{align}
Similar to the sample-wise consistency, we compute the consistency between one network's one view prediction with the other network's the other view prediction, leading to two cross-correlation matrices. As done above, we convert $\mathbf{\mathbf{R}}_1$ and $\mathbf{\mathbf{R}}_2$ to symmetric ones and normalize its rows and columns:
\begin{align}
\mathbf{\mathbf{\hat{R}}}_i = \phi\big((\mathbf{\mathbf{R}}_i + \mathbf{\mathbf{R}}_i^\top) / 2\big),~i=1,2.
\end{align}
Then, we have our class-wise cross consistency loss as,
\begin{align}
\mathcal{L}_{class}^{cross} =  -\frac{1}{2C}\big(\Tr({\mathbf{\hat{R}}}_1) + \Tr(\mathbf{\mathbf{\hat{R}}}_2)\big),
\label{eq:class_cross}
\end{align}

\noindent \textbf{Class-wise Consistency Loss.}
Similar to the sample-wise one, our final class-wise cross-consistency loss is:
\begin{equation}
\begin{aligned}
\mathcal{L}_{class} &= \frac{1}{2}(\mathcal{L}_{class}^{inner} + \mathcal{L}_{class}^{cross}).
\label{eq:class_v1}
\end{aligned}
\end{equation}

\noindent \textbf{Analysis.}
We also devise a simple analysis of class-wise consistency, showing that it integrates all the merits claimed by Mutual Information Maximization~\cite{shot}, Minimum Class Confusion~\cite{jin2020minimum}, and sample-wise consistency. This is achieved by simultaneously addressing three objectives: i) \textbf{consistency regularization}, ensuring that predictions derived from distinct augmented views remain consistent, \ie, $\mathbf{P}^{\prime} = \mathbf{P}^{\dprime}$; ii) \textbf{entropy minimization}, facilitating the sharpening of softmax predictions for individual samples, \ie, the rows of $\mathbf{P}$ are one-hot-like; and iii) \textbf{prediction diversification}, penalizing collapsed solutions in which the majority of samples are predicted to a single class, \ie, the rows of $\mathbf{P}$ are distinct.
Specifically, without loss of generality, assuming that the batch prediction is a square matrix, it can be verified that the optimal solutions of the class-wise consistency are equal to the ones of the following optimization problem:
\begin{align}
& \min_{\mathbf{P}_1, \mathbf{P}_2} \Vert\frac{1}{2} (\mathbf{P}_1^\top\mathbf{P}_2 + \mathbf{P}_2^\top\mathbf{P}_1)  - \mathbf{I}_n\Vert_F, \\
& \mathrm{s.t.} \quad \mathbf{P}_i\mathbf{1}_n = \mathbf{1}_n,~\mathbf{P}_i \ge 0,~i = 1,2.
\end{align}
The optimal solution for this convex optimization problem is $\mathbf{P}_1=\mathbf{P}_2$, which are permutation matrices that achieve the three aforementioned objectives. In addition, our class-wise cross consistency further benefits the co-learning of the two networks.

\subsection{Overall Objective Function}
Both the proposed $\mathcal{L}_{sample}$ (Eq.~\ref{eq:sample_v1}) and $\mathcal{L}_{class}$ (Eq.~\ref{eq:class_v1}) are designed for the unlabeled sample $\mathcal{D}_t^u$, while for the labeled samples in $\mathcal{D}_t^l$, we employ the cross entropy loss:
\begin{align}
    \mathcal{L}_{CE} = -\sum_{(\mathbf{x},\mathbf{y}) \sim \mathcal{D}_t^l} \mathbf{y}[\log\big( F(\mathbf{x}|\theta_s) \big) + \log\big( F(\mathbf{x}|\theta_t) \big)].
    \label{eq:ce}
\end{align}
Summing up the three loss functions, we have our overall objective:
\begin{align}
    \min_{\theta_s, \theta_t} \mathcal{L}_{CE} + \lambda_1 \mathcal{L}_{sample} + \lambda_2 \mathcal{L}_{class},
    \label{eq:overall}
\end{align}
where $\lambda_1$ and $\lambda_2$ are hyper-parameters control the trade-off between losses.

\vspace*{0.5mm} \noindent \textbf{Choice of Final Model.}
We observe that the performance of $F(\cdot|\theta_s)$ and $F(\cdot|\theta_t)$ converge to a similar point after training. 
Therefore, without loss of generality, we choose $F(\cdot|\theta_s)$ as the final model for inference.

\section{Experiments}

\begin{table*}[t]
    \centering    \caption{\textit{H-score} on \textit{DomainNet} under the settings of 3-shot and 5-shot using ResNet34 as backbone networks. }
    \label{res_domainnet}
    \vspace{-2mm}
    \setlength\tabcolsep{8.8pt}
    \scalebox{0.7}{
        \begin{tabular}{l|cccccccccccccc}
        \toprule
        & \multicolumn{2}{c}{R $\rightarrow$ C} 
        & \multicolumn{2}{c}{P $\rightarrow$ C} 
        & \multicolumn{2}{c}{C $\rightarrow$ S}
        & \multicolumn{2}{c}{R $\rightarrow$ P} 
        & \multicolumn{2}{c}{S $\rightarrow$ P}
        & \multicolumn{2}{c}{R $\rightarrow$ S} 
        % & \multicolumn{2}{c|}{P $\rightarrow$ R}
        & \multicolumn{2}{c}{\textbf{Mean}} \\  
        \textbf{Method}
        & \small 3-shot & \small 5-shot
        & \small 3-shot & \small 5-shot
        & \small 3-shot & \small 5-shot
        & \small 3-shot & \small 5-shot
        & \small 3-shot & \small 5-shot
        & \small 3-shot & \small 5-shot
        % & \small 1-shot & \small 10-shot
        & \small \textbf{3-shot} & \small \textbf{5-shot}\\
        \midrule
        CE     & 50.8 & 58.0 & 53.0 & 57.3 & 47.1 & 54.6 & 50.3 & 57.7 & 53.5 & 58.8 & 43.6 & 50.6 & 49.7 & 56.2   \\
        % ENT   & 45.9 & 62.0 & 44.3 & 56.1 & 36.0 & 57.4 & 36.8 & 51.5 & 43.0 & 59.2 & 36.3 & 44.6 & 40.4 & 55.1  \\
        MixMatch & 48.0 & 66.5 & 50.5 & 65.2 & 44.4 & 57.6 & 50.1 & 64.6 & 50.2 & 65.3 & 44.2 & 58.1 & 47.9 & 62.8  \\
        FixMatch & 52.5 & 68.6 & 48.2 & 66.4 & 49.8 & 61.1 & 52.8 & 68.5 & 48.6 & 68.2 & 43.0 & 64.2 & 49.2 & 66.2  \\
        UMA  & 42.7 & 60.4 & 46.2 & 58.5 & 41.7 & 51.2 & 42.9 & 56.8 & 40.4 & 51.6 & 40.2 & 50.5 & 42.3 & 54.8  \\
        MME  & 56.7 & 66.5 & 58.9 & 64.1 & 45.0 & 55.2 & 49.9 & 62.4 & 52.7 & 63.8 & 42.8 & 51.2 & 51.0 & 60.5   \\
        CDAC & 59.0 & 71.4 & 59.2 & 72.3 & 50.8 & 64.7 & 62.3 & 70.1 & 62.3 & 70.5 & 51.9 & 63.1 & 57.6 & 68.7  \\
        % SSHT & 64.1 & 73.1 & 66.8 & 72.2 & 54.2 & 61.0 & 59.0 & 66.0 & 65.6 & 67.0 & 47.1 & 57.7 & 59.5 & 66.2   \\
        MCL & 62.0 & 74.8 & 65.2 & 73.5 & 53.4 & 65.1 & 60.9 & 72.5 & 64.7 & 70.9 & 54.9 & 63.1 & 60.2 & 70.0   \\
        AdaMatch & 59.5&73.3&62.1&71.4&52.0&64.7&60.5&73.1&65.9&73.1&51.6&60.8&58.6&69.4  \\
        SHOT$++$ & 55.6 & 66.8 & 59.4 & 68.1 & 51.0 & 58.5 & 56.1 & 63.3 & 56.7 & 60.8 & 45.4 & 53.6 & 54.0 & 61.9  \\
        \midrule
        CCT & \textbf{69.9} & \textbf{77.7} & \textbf{69.0} & \textbf{77.4} & \textbf{58.6} & \textbf{66.8} & \textbf{66.7} & \textbf{75.5} & \textbf{67.4} & \textbf{75.3} & \textbf{56.2} & \textbf{66.9} & \textbf{64.6} & \textbf{73.3}  \\
        \bottomrule
        \end{tabular}
    }
    \label{table:domainnet}
    % \vspace{-2mm}
\end{table*}

\begin{table*}[t]
    \centering
    \caption{\textit{H-score} on \textit{Office-Home} under the settings of 3-shot and 5-shot using ResNet34 as backbone networks. }
    \label{res_officehome}
    \vspace{-2mm}
    \setlength\tabcolsep{9pt}
    \scalebox{0.7}{
        \begin{tabular}{l|ccccccccccccc}
        \toprule
        \textbf{Method}
        & \small A $\rightarrow$ C
        & \small A $\rightarrow$ P
        & \small A $\rightarrow$ R
        & \small C $\rightarrow$ A
        & \small C $\rightarrow$ P
        & \small C $\rightarrow$ R
        & \small P $\rightarrow$ A
        & \small P $\rightarrow$ C
        & \small P $\rightarrow$ R
        & \small R $\rightarrow$ A
        & \small R $\rightarrow$ C
        & \small R $\rightarrow$ P
        & \textbf{Mean} \\  
        \midrule
        \multicolumn{14}{c}{\textbf{3-shot}}\\
        \midrule
        CE       & 51.8 & 73.9 & 69.9 & 49.9 & 70.7 & 65.6 & 57.8 & 51.1 & 68.9 & 55.7 & 54.0 & 74.6 & 62.0 \\
        % ENT      & 53.2 & 76.4 & 71.7 & 46.3 & 74.9 & 66.1 & 54.6 & 46.5 & 69.4 & 57.3 & 54.8 & 77.2 & 62.4 \\
        MixMatch & 51.3 & 77.6 & 73.3 & 51.8 & 76.4 & 69.7 & 56.8 & 51.1 & 72.2 & 58.3 & 53.4 & 78.5 & 64.2 \\
        FixMatch & 50.6 & 76.2 & 67.5 & 39.9 & 76.6 & 66.9 & 53.9 & 56.7 & 67.4 & 47.1 & 51.8 & 75.4 & 60.8 \\
        UMA     & 54.6 & 76.3 & 71.3 & 52.1 & 73.7 & 63.3 & 57.2 & 52.5 & 68.8 & 54.0 & 56.4 & 76.2 & 63.0 \\
        MME      & 55.6 & 77.1 & 72.3 & 55.2 & 75.3 & 68.8 & 60.0 & 53.6 & 72.5 & 61.4 & 57.8 & 77.3 & 65.6 \\
        CDAC     & 55.7 & 76.3 & 72.0 & 54.0 & 75.4 & 68.9 & 54.5 & 58.8 & 72.3 & 60.6 & 57.1 & 76.9 & 65.2 \\
        % SSHT     & 55.0 & 73.0 & 69.9 & 56.0 & 72.3 & 67.0 & 51.4 & 53.9 & 68.6 & 58.0 & 59.1 & 74.0 & 63.2 \\
        MCL     & 54.2&74.2&71.6&54.6&77.6&66.6&54.5&59.6&73.4&60.0&55.8&77.3 & 65.0 \\
        AdaMatch 
 & 52.2&71.4&71.1&55.4&78.6&65.4&55.5&59.7&74.9&59.2&54.0&77.8 & 64.5 \\
        SHOT++   & 54.9 & 80.3 & 74.2 & 53.3 & 73.9 & 70.8 & 55.9 & 52.3 & 76.9 & 50.2 & 52.7 & 79.3 & 64.6 \\
        \midrule
        CCT     & \textbf{57.8} & \textbf{80.6} & \textbf{77.3} & \textbf{61.0} & \textbf{79.4} & \textbf{76.9} & \textbf{62.6} & \textbf{57.8} & \textbf{77.1} & \textbf{63.4} & \textbf{59.2} & \textbf{80.2} & \textbf{69.4} \\
        \midrule[0.7pt]
        \multicolumn{14}{c}{\textbf{5-shot}}\\
        \midrule
        CE    &  56.5 & 77.1 & 73.2  & 58.7 & 75.2 & 69.4 & 61.4 & 56.5 & 73.2 &    60.6 & 57.2 & 78.9 & 66.5 \\
        % ENT      & 61.4 & 80.5 & 77.3 & 61.4 & 80.2 & 73.9 & 62.2 & 59.6 & 78.5 & 63.9 & 61.8 & 83.2 & 70.3 \\
        MixMatch & 59.1 & 81.5 & 76.7 & 61.3 & 80.2 & 75.3 & 62.6 & 59.9 & 77.3 & 62.6 & 58.5 & 83.6 & 70.2 \\
        FixMatch    & 59.1 & 80.8 & 76.7 & 58.5 & 79.9 & 72.4 & 60.3 & 61.6 & 78.0 & 61.7 & 62.5 & 80.0 & 69.3 \\
        UMA & 55.8 & 77.6 & 72.3 & 56.4 & 76.7 & 68.5 & 60.0 & 59.9 & 75.3 & 55.6 & 52.4 & 77.7 & 65.7 \\
        MME      & 62.2 & 80.5 & 74.3 & 63.1 & 81.6 & 73.4 & 62.6 & 61.9 & 76.5 & 64.0 & 62.4 & 83.1 & 70.5 \\
        CDAC & 64.0 & 80.6 & 76.3 & 61.5 & 80.2 & 71.8 & 60.1 & 62.5 & 76.1 & 61.4 & 59.7 & 78.3 & 69.4 \\
        % SSHT & 57.7 & 79.5 & 71.6 & 58.7 & 79.6 & 68.1 & 56.9 & 61.5 & 72.2 & 54.9 & 59.5 & 78.7 & 66.6 \\
        MCL & 63.5&79.5&76.9&63.1&80.4&73.5&61.1&64.3&78.2&61.8&59.4&79.7 & 70.1 \\
        AdaMatch   & 57.6&78.7&76.3&59.1&82.1&70.1&60.3&63.4&78.1&61.1&61.2&80.4
& 69.0 \\
        SHOT++ & 54.6 & 80.3 & 75.2 & 62.2 & 75.0 & 74.1 & 62.9 & 55.5 & 77.5 & 64.1 & 50.9 & 80.6 & 67.7 \\
        \midrule
        CCT     & \textbf{65.9} & \textbf{82.2} & \textbf{79.3} & \textbf{67.3} & \textbf{82.1} & \textbf{77.9} & \textbf{67.9} & \textbf{65.6} & \textbf{78.9} & \textbf{67.5} & \textbf{63.6} & \textbf{83.9} & \textbf{73.5} \\
        % \midrule[0.7pt]
        % \multicolumn{14}{c}{\textbf{10-shot}}\\
        % \midrule
        % CE  & 66.5 & 83.9 & 77.8 & 68.3 & 81.9 & 76.5 & 68.1 & 66.4 & 79.1 &    68.5 & 65.6 & 83.5 & 73.8 \\
        % ENT      & 70.4 & 87.0 & 81.4 & 71.8 & 85.2 & 79.6 & 71.7 & 70.6 & 82.5 & 73.2 & 70.2 & 86.9 & 77.5 \\
        % MixMatch & 72.0 & 86.9 &  81.4  & 69.8 & 86.6 & 81.7 & 68.1 & 71.1 & 82.6 & 70.2 & 73.1 & 86.9 & 77.5 \\
        % FixMatch    & 71.5 & 86.1 & 80.0 & 69.9 & 85.0 & 79.3 & 69.7 & 72.1 & 82.5 & 69.8 & 72.0 & 86.3 & 77.0 \\
        % UMA & 69.4 & 84.3 & 80.8 & 64.3 & 81.1 & 79.5 & 69.3 & 66.7 & 78.3 & 68.7 & 68.3 & 80.0 & 74.2  \\
        % MME      & 71.5 & 86.4 & 80.8 & 71.4 & 85.6 & 79.1 & 70.8 & 71.0 & 81.3 & 71.9 & 71.5 & 87.0 & 77.4 \\
        % CDAC & 72.5 & 85.5 & 80.6 & 65.6 & 83.6 & 78.9 & 68.5 & 71.5 & 80.1 & 69.3 & 73.0 & 86.2 & 76.3 \\
        % SSHT & 71.9 & 84.3 & 76.2 & 63.9 & 83.8 & 74.6 & 65.1 & 67.6 & 77.1 & 63.7 & 70.6 & 83.0 & 73.5 \\
        % SHOT++ & 66.7 & 84.9 & 79.6 & 66.6 & 83.6 & 78.6 & 67.9 & 64.5 & 79.3 & 68.2 & 66.3 & 83.0 & 74.2 \\
        % \midrule
        % CCT     & \textbf{74.5} & \textbf{87.2} & \textbf{82.8} & 70.4 & \textbf{87.3} & \textbf{81.9} & 70.6 & \textbf{74.1} & \textbf{82.7} & 71.7 & \textbf{74.6} & \textbf{87.1} & \textbf{78.7} \\
        \bottomrule
        \end{tabular}
    }
    \label{table:office_home}
    % \vspace{-2mm}
\end{table*}

\begin{table}[t]
    \centering
    \caption{\textit{H-score} on \textit{Office} under the settings of 5-shot and 10-shot using ResNet34 as backbone networks. }
    \label{table:office}
    % \vspace{2mm}
    \setlength\tabcolsep{10pt}
    \scalebox{0.7}{
        \begin{tabular}{l|cccc}
        \toprule
        & \multicolumn{2}{c}{W $\rightarrow$ A} 
        & \multicolumn{2}{c}{D $\rightarrow$ A} \\
        \textbf{Method}
        & \small 3-shot & \small 5-shot
        & \small 3-shot & \small 5-shot \\
        \midrule
        CE       & 60.4 & 61.2 & 60.9 & 61.6   \\
        % ENT      & 63.6 & 67.3 & 61.1 & 66.7   \\
        Mixmatch & 61.2 & 69.7 & 62.6 & 69.4  \\
        Fixmatch & 62.4 & 63.4 & 61.0 & 67.4   \\
        UMA      & 60.1 & 61.5 & 58.7 & 62.4   \\
        MME      & 64.7 & 69.0 & 64.3 & 69.1   \\
        CDAC     & 64.5 & 68.5 & 63.9 & 67.8   \\
        MCL     & 65.9 & 68.8 & 66.0 & 68.1  \\
        AdaMatch & 62.8 & 65.4 & 64.1 & 67.4   \\
        SHOT++   & 64.0 & 68.3 & 64.5 & 68.2   \\
        \midrule
        CCT      & \textbf{68.3} & \textbf{71.3} & \textbf{67.9} & \textbf{70.7}   \\
        \bottomrule
        \end{tabular}
    }
    \vspace{-2mm}
\end{table}

\subsection{Experimental Setup}
% \vspace{1mm}
\noindent\textbf{Datasets.}
We evaluate our method on several popular benchmark datasets, including \textit{Office}~\cite{saenko2010adapting}, \textit{DomainNet}~\cite{peng2019domainnet}, and \textit{Office-Home}~\cite{venkateswara2017officehome}, with different $\mathcal{C}_s$, $\mathcal{C}_t$ and $\xi$ (see Sec.~\ref{sec:problem_definition}). Similar to most recent works~\cite{yang2020mico,li2021cross}, we conduct 3-shot (3 labeled samples per class in the target domain) and 5-shot experiments on all datasets.
\noindent \textbf{DomainNet}~\cite{peng2019domainnet} was first introduced as a multi-source domain adaptation benchmark comprising 6 domains with 345 categories. Following~\cite{saito2019semi,li2021cross}, we select the \textit{Real}, \textit{Clipart}, \textit{Painting}, \textit{Sketch} domains with 126 categories for evaluation. The first 80 classes are used as $\mathcal{C}_s$ and last 96 classes are used as $\mathcal{C}_t$, hence $|\mathcal{C}| = 50$ and $\xi = 0.4$.
\noindent \textbf{Office-Home}~\cite{venkateswara2017officehome} is a popular domain adaptation benchmark, which consists of 4 domains (\textit{Real}, \textit{Clipart}, \textit{Product}, \textit{Art}) and 65 categories. We use the first 43 classes as $\mathcal{C}_s$ and the last 35 classes as $\mathcal{C}_t$, hence $|\mathcal{C}| = 13$ and $\xi = 0.2$.
\noindent \textbf{Office}~\cite{saenko2010adapting} contains 3 domains (Amazon, Webcam, and DSLR) with 31 classes. Similar to~\cite{saito2019semi,li2021cross}, we conduct experiments from DSLR to Amazon and Webcom to Amazon to evaluate on the domain with enough examples. The first 25 classes are set as $\mathcal{C}_s$ and the last 25 classes are set as $\mathcal{C}_t$, hence $|\mathcal{C}|$ = 19 and $\xi = 0.6$.

% \vspace{1mm}
\noindent\textbf{Evaluation Criterion.}
Following~\cite{fu2020learning}, we use \textit{H-score} to evaluate the performance of our method, which is defined as:
\begin{equation}
H = 2 \cdot \frac{a_{\mathcal{C}} \cdot a_{\mathcal{\Bar{C}}_t}}{a_{\mathcal{C}} + a_{\mathcal{\Bar{C}}_t}},
\end{equation}
where $a_{\mathcal{C}}$ and $a_{\mathcal{\Bar{C}}_t}$ represent the accuracy for the common class $\mathcal{C}_t$ and the target private class $\mathcal{\Bar{C}}_t$, respectively. 

\begin{table*}[ht]
    \centering
    \caption{\textit{H-score} on \textit{Domainnet} under the settings of 5-shot using ResNet34. The first row means only $\mathcal{L}_{CE}$~(Eq.~\ref{eq:ce}) is utilized.}
    % \vspace{2mm}
    \label{res_ablation}
    \setlength\tabcolsep{10pt}
    \scalebox{0.7}{
        \begin{tabular}{ccccc|ccccccc}
        \toprule
        & $L_{sample}^{inner}$
        & $L_{class}^{inner}$
        & $L_{sample}^{cross}$
        & $L_{class}^{cross}$
        &  R $\rightarrow$ C
        &  P $\rightarrow$ C
        &  C $\rightarrow$ S
        &  R $\rightarrow$ P
        &  S $\rightarrow$ P
        &  R $\rightarrow$ S
        & \textbf{Mean} \\  
        \midrule
         ($a$) & & & &  & 58.0  & 57.3  & 54.6  & 57.7  & 58.8  & 50.6  & 56.2  \\
        ($b$) & \checkmark & & & & 68.6 & 66.4 & 61.1 & 65.5 & 68.2 & 64.2 & 65.7  \\
        ($c$) & \checkmark & & \checkmark & & 71.8 & 67.9 & 65.2 & 70.4 & 70.9 & 64.7 & 68.5 \\
        ($d$) & \checkmark & \checkmark & & & 76.2 & 75.8 & 66.3 & 74.4 & 72.9 & 64.8 & 71.7  \\
        ($e$) & & & \checkmark & \checkmark & 77.6 & 77.2 & 67.2 & 73.8 & 74.8 & 66.0 & 72.8  \\
        ($f$) & \checkmark & \checkmark & \checkmark & \checkmark & 77.7 & 77.4 & 66.8 & 75.5 & 75.3 & 66.8 & 73.3  \\
        \bottomrule
        \end{tabular}
    }
    \label{table:ablation}
    % \vspace{2mm}
\end{table*}

% \begin{figure*}[ht]
% 	\begin{subfigure}{.33\textwidth}
% 		\centering
% 		\includegraphics[width=.99\linewidth]{latex/Figures/private_size.png}
% % 		\captionsetup{labelformat=empty}
% 		\subcaption{H-score w.r.t. $|\mathcal{\Bar{C}}_t|$}
% 		\label{fig:p_size}
% 	\end{subfigure}%
% 	\begin{subfigure}{.33\textwidth}
% 		\centering
% 		\includegraphics[width=.99\linewidth]{latex/Figures/common_size.png}
% % 		\captionsetup{labelformat=empty}
% 		\subcaption{H-score w.r.t. $|\mathcal{C}|$}
% 		\label{fig:c_size}
% 	\end{subfigure}%
% 	\begin{subfigure}{.33\textwidth}
% 		\centering
% 		\includegraphics[width=.93\linewidth]{latex/Figures/numshot.png}
% % 		\captionsetup{labelformat=empty}
% 		\subcaption{H-score w.r.t. $|\mathcal{D}_t^l|/C$}
% 		\label{fig:num_shot}
% 	\end{subfigure}%
% 	\caption{(a) H-score w.r.t. $|\mathcal{\Bar{C}}_t|$ in task R $\rightarrow$ C of \textit{Office-Home} with $\xi=0.23$. (b) H-score w.r.t. $|\mathcal{C}|$ in task R $\rightarrow$ C of \textit{Office-Home}. (c) H-score w.r.t. number of labeled samples per-class in task R $\rightarrow$ C of \textit{Domainnet}.}
% % 	\label{fig:vis_analysis}
% \end{figure*}
\begin{figure*}[t]
    \centering
    \begin{minipage}{0.3\textwidth}
        % \label{fig:p_size}
        \centering
        \includegraphics[width=.99\linewidth]{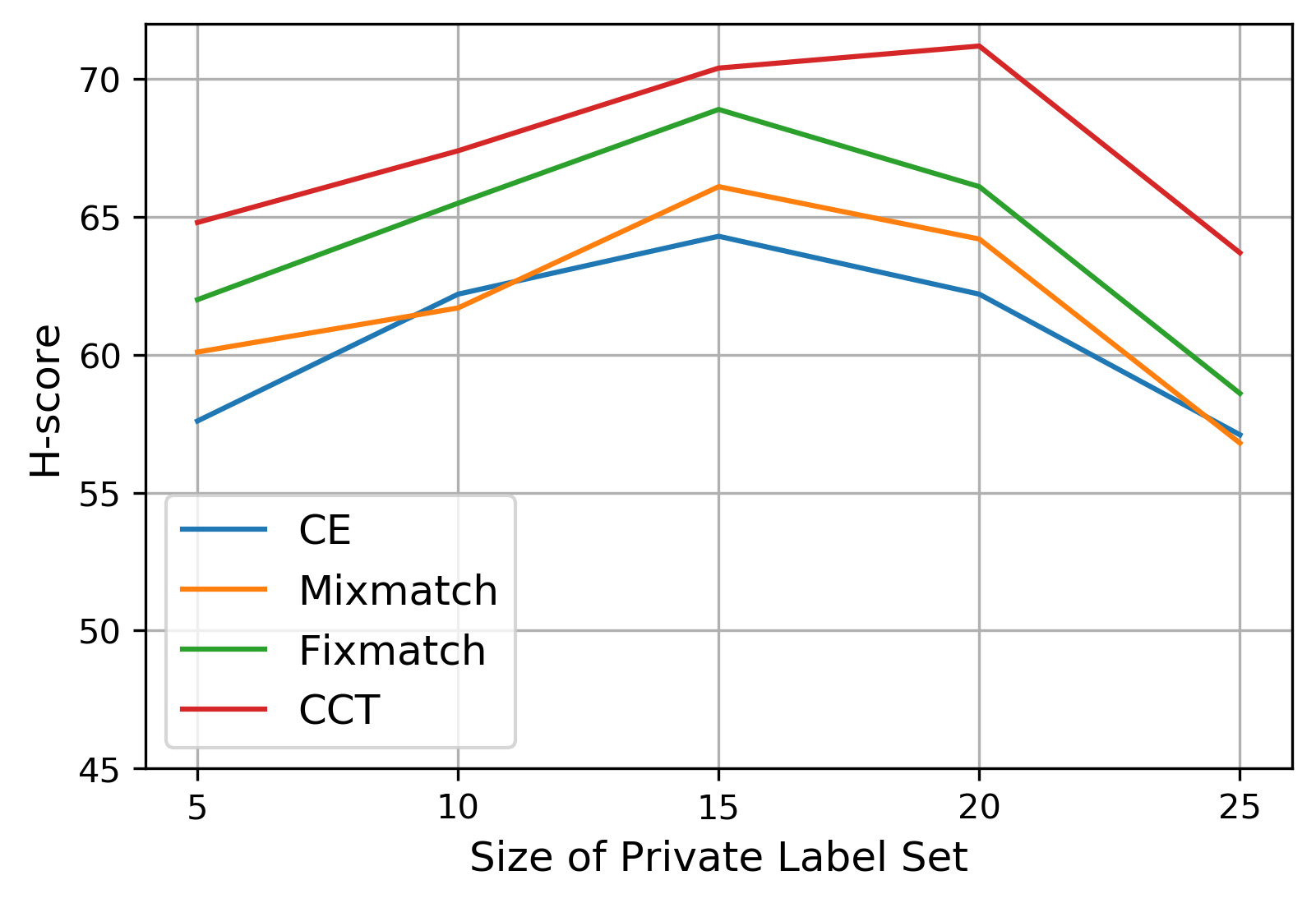}\\
        (a) \textit{H-score} w.r.t. $|\mathcal{\Bar{C}}_t|$.
    \end{minipage}
    \begin{minipage}{.3\textwidth}
        \centering
        \includegraphics[width=.99\linewidth]{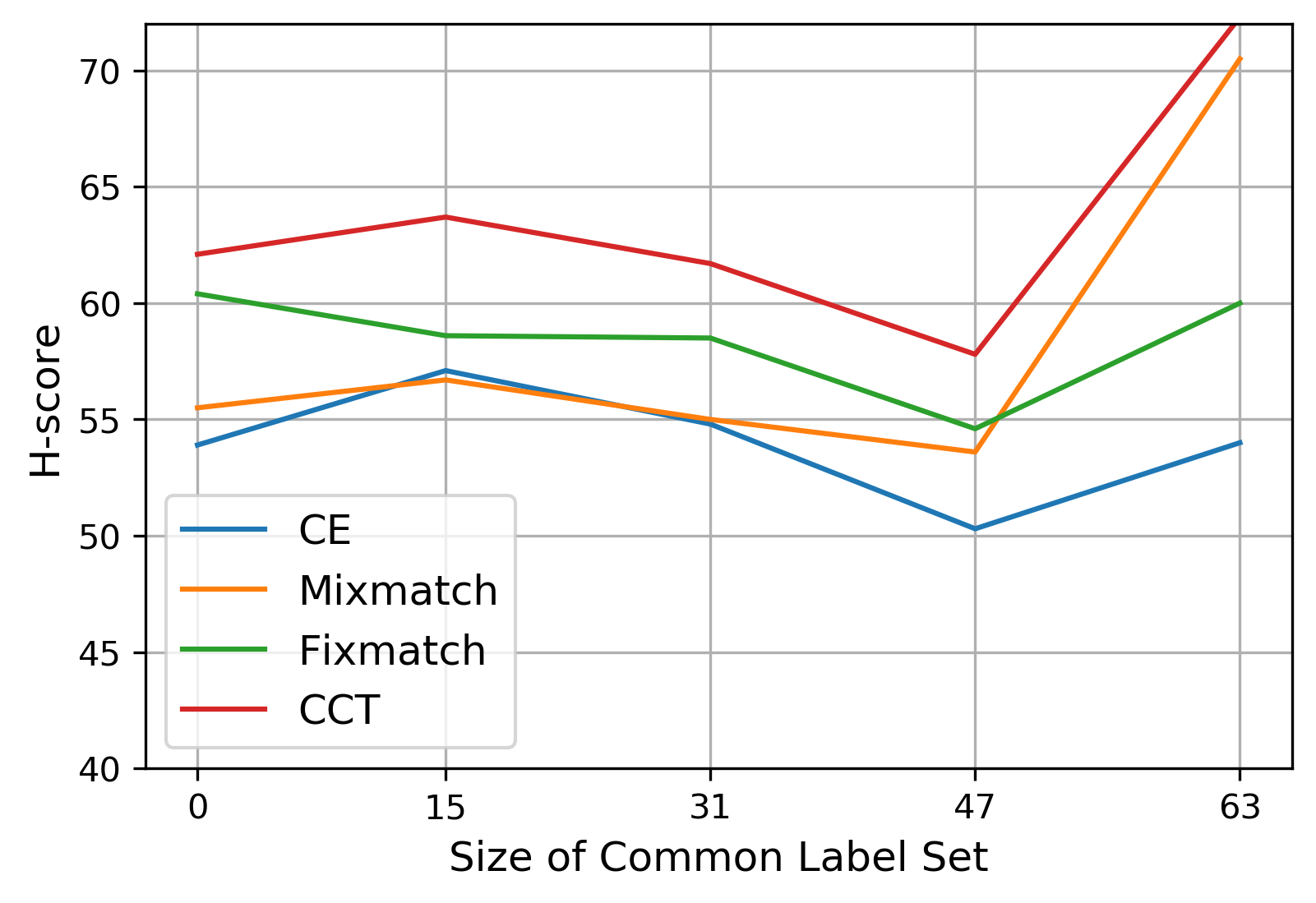}\\
        (b) \textit{H-score} w.r.t. $|\mathcal{C}|$
        % \label{fig:c_size}
        \end{minipage}%
    \begin{minipage}{.3\textwidth}
    \centering
        \includegraphics[width=.93\linewidth]{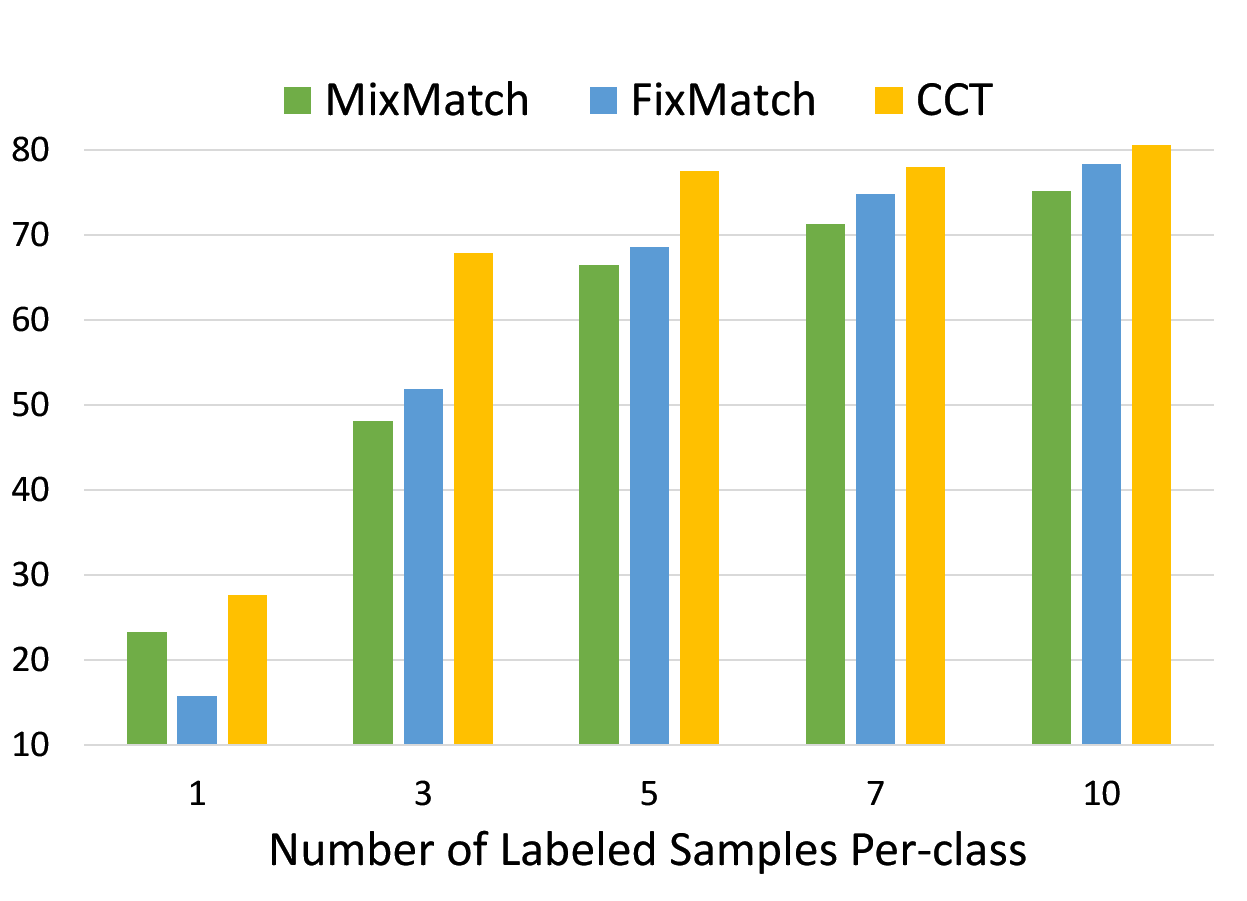}\\
        (c) \textit{H-score} w.r.t. $|\mathcal{D}_t^l|/C$
        % \label{fig:num_shot}
    \end{minipage}%
    \caption{(a) \textit{H-score} w.r.t. $|\mathcal{\Bar{C}}_t|$ in task R $\rightarrow$ C of \textit{Office-Home} with $\xi=0.23$. (b) \textit{H-score} w.r.t. $|\mathcal{C}|$ in task R $\rightarrow$ C of \textit{Office-Home}. (c) \textit{H-score} w.r.t. number of labeled samples per-class in task R $\rightarrow$ C of \textit{Domainnet}.}
    \label{fig:multitable}
    \vspace{-2mm}
\end{figure*}

\vspace{1mm}
\noindent\textbf{Implementation Details.}
Similar to~\cite{saito2019semi,li2021cross}, we use Resnet34~\cite{resnet} as the backbone network where the classifier consists of a $L_2$ normalization layer that projects feature into a spherical feature space and a linear layer. Following~\cite{shot}, we pre-train the source model with label smoothing~\cite{muller2019does} for 50 epochs. For the pre-training of the target-only network, we train the network for 50 epochs with supervised contrastive loss~\cite{khosla2020scl} for the labeled samples and SimCLR~\cite{chen2020simclr} loss for the unlabeled samples. Moreover, we use RandAugment~\cite{cubuk2020randaugment} as the augmentation for $x^{\dprime}$ and standard random resize, flip and crop as the augmentation for $x^{\prime}$. For the adaptation phase, the experimental settings (\ie, optimizer, batch size, \etc) are the same as MME~\cite{saito2019semi}. We set $\lambda_1$ as 1, and $\lambda_2$ as 1 for \textit{Domainnet} and 0.5 for \textit{Office} and \textit{Office-Home}. All experiments are implemented with PyTorch~\cite{pytorch} on a single NVIDIA 2080Ti.

\subsection{Comparison Experiments}

We compared our method with i) baseline methods: only train the labeled target samples with cross-entropy loss (CE), ii) semi-supervised domain adaptation (SSDA) methods: MME\cite{saito2019semi}, CDAC~\cite{li2021cross}, MCL~\cite{yan2022multi} and AdaMatch~\cite{berthelot2022adamatch}, iii) semi-supervised learning (SSL) methods: MixMatch~\cite{berthelot2019mixmatch} and FixMatch~\cite{sohn2020fixmatch}, iv) a semi-supervised model adaptation (SSMA) method: SHOT++~\cite{shotplus}, and v) a universal model adaptation method: UMA~\cite{kundu2020universal}.
To make a fair comparison, we implement them with the same $F(\cdot|\mathbf{\theta}_s)$ as used in our method. Note that we only report the results using $F(\cdot|\mathbf{\theta}_s)$ as it outperforms those using $F(\cdot|\mathbf{\theta}_t)$.
The implementation details of the compared methods can be found in the supplementary file.
The results on the three benchmark datasets are shown in Table~\ref{res_domainnet},~\ref{res_officehome} and ~\ref{table:office}, respectively. 
It can be observed that our method significantly outperforms the state-of-the-art SSL, SSDA, SSMA, and UMA methods in terms of H-score.

% 
% \vspace{2mm}
\noindent\textbf{Results.}
% \subsubsection{Results.}
As Table~\ref{res_domainnet} shows, our method outperforms all previous methods in all domains on \textit{Domainnet}, achieving 64.6\% and 73.3\% mean H-scores under 3-shot and 5-shot settings, respectively. Notably, our method surpasses the State-of-the-Art (SOTA) SSDA technique, \ie, AdaMatch~\cite{berthelot2022adamatch}, by substantial margins, demonstrating the superiority of the proposed method. 
Furthermore, it is imperative to highlight that certain methods falter even in the 3-shot setting, exhibiting performance inferior to the basic CE method. In contrast, CCT consistently maintains its stability and superiority.
As Table~\ref{res_officehome} and~\ref{table:office} show, CCT still works effectively and outperforms all the previous methods on \textit{Office-Home} and \textit{Office} significantly, which further demonstrates the efficacy of the proposed method.

% \vspace{2mm}
% \noindent\textbf{Ablation Study.}
\subsection{Ablation Study}
We perform an ablation study on \textit{Domainnet} to verify the efficacy of each component of the proposed method. For clarity, except for the most basic sample-wise inner consistency, either sample/class-wise or inner/cross consistency are jointly used. 
% The results are shown in Table~\ref{table:ablation}.
As Table~\ref{table:ablation} shows, starting from $\mathcal{L}_{sample}^{inner}$ (b) that represents the case when only the most basic sample-wise inner consistency is used, adding the sample-wise cross consistency loss~(c) improves the performance by 2.8\% and a further incorporation of the class-wise consistency losses~(f) boosts the performance by 4.7\%. 
This demonstrates the efficacy of both the proposed collaborating consistency training framework, especially the cross consistency, and the proposed class-wise consistency regularization. 
In general, our final method outperforms its variant with only the basic sample-wise inner consistency regularization by 7.6\%. 
Surprisingly, we observed that the cross consistency regularization could work effectively without the inner consistency regularization~(e), and performs even better than inner consistency alone~(d) by 1.1\%. We conjecture that the inner consistency can be \textit{implicitly} regularized by the cross consistency, and they are compatible as the network achieves better performance when they are combined together.
Furthermore, the 6\% improvement of sample and class-wise inner consistency (d) over sample-wise inner consistency alone (b) demonstrates the efficacy and versatility of the proposed class-wise consistency.

\subsection{Analysis Experiments}
In this section, we conduct several additional experiments to validate the versatility of our method. Further analysis experiments can be found in the supplementary file.

\vspace{1mm}
\noindent \textbf{Varying Sizes of $\mathcal{\Bar{C}}_t$ and $\mathcal{\Bar{C}}_s$.} 
Similar to~\cite{you2019universal}, with fixed $|\mathcal{C}_s \cup \mathcal{C}_t|$ and $\xi$, we investigate how the performance of our method changes with various sizes of $\mathcal{\Bar{C}}_t$.
We conduct the experiments on the task of \textit{Real} $\rightarrow$ \textit{Clipart} of \textit{Office-Home}. We set $\xi$ as 0.23 and $\mathcal{C} = 15$. Note that $|\mathcal{\Bar{C}}_s|$ will change correspondingly with  $|\mathcal{\Bar{C}}_t|$. 
As Fig.~\ref{fig:multitable}a shows, our method outperforms all the other methods for all $|\mathcal{\Bar{C}}_t|$. Furthermore, compared with MixMatch~\cite{berthelot2019mixmatch} and FixMatch~\cite{sohn2020fixmatch}, the performance gap increases as the size of $|\mathcal{\Bar{C}}_t|$ increases, which implies that our method (CCT) is more suitable for the scenarios with large target-private categories.

\vspace{1mm}
\noindent \textbf{Varying Sizes of $\mathcal{C}$.}
% \subsubsection{Varying Sizes of $\mathcal{C}$}
Similarly, we investigate how the performance of our method changes with various sizes of $\mathcal{C}$ but fixed $|\mathcal{C}_s \cup \mathcal{C}_t|$. We conduct the experiments on the task of \textit{Real} $\rightarrow$ \textit{Clipart} of \textit{Office-Home}. We set $|\mathcal{C}_s \cup \mathcal{C}_t| = 65$ and $|\mathcal{\Bar{C}}_t| = |\mathcal{\Bar{C}}_s|$. The results are shown in Fig.~\ref{fig:multitable}b. Note that when $|\mathcal{C}| = 0$, we set $\mathcal{\Bar{C}}_s = \mathcal{\Bar{C}}_t = 15$ and report the accuracy of target-private categories samples only. It can be observed that our method (CCT) outperforms all the other methods for all $|\mathcal{C}|$. It is worth noting that our CCT can still achieve decent performance when there is no intersection between the source and target label sets.

\vspace{1mm}
\noindent \textbf{Varying Sizes of $\mathcal{D}_t^l$.}
% \subsubsection{Varying Sizes of $\mathcal{D}_t^l$}
Although we already show the superior performance of our method in the 3-shot and 5-shot settings, it is still interesting to see how the performance of our method changes under various number of labeled samples. 
We conduct the experiments on the task of \textit{Real} $\rightarrow$ \textit{Clipart} of \textit{Domainnet}. As Fig.~\ref{fig:multitable}c shows, our method (CCT) consistently outperforms MixMatch~\cite{berthelot2019mixmatch} and FixMatch~\cite{sohn2020fixmatch} in all cases. Specifically, when the number of labeled samples is small, CCT surpasses the other methods.

% \begin{table}[t]
%     \centering
%     % \vspace{-2mm}
%     % \caption{}
%     % \vspace{-3mm}
%     \caption{SSMA Results on \textit{Office-Home} 1-shot setting using VGG16 as the backbone.}
%     \vspace{-2mm}
%     \setlength\tabcolsep{10pt}
%     \scalebox{0.8}{
%         \begin{tabular}{l|c}
%         \toprule
%         \textbf{Method}
%         & \textbf{Mean Accuracy} \\
%         \midrule
%         CE & 57.4 \\
%         MME & 62.7 \\
%         SHOT-IM$++$ & 65.2 \\
%         SHOT$++$ & 66.1 \\
%         CCT & \textbf{68.3} \\
%         \bottomrule
%         \end{tabular}
%     }
%     \label{table:ssma}
%     % \label{table:L2loss}
%     \vspace{-3mm}
% \end{table}
\begin{table}[t]
    \centering
    % \vspace{-2mm}
    % \caption{}
    % \vspace{-3mm}
    \caption{SSMA Results on \textit{Office-Home} 1-shot setting using VGG16 as the backbone.}
    % \vspace{-2mm}
    \setlength\tabcolsep{6pt}
    \scalebox{0.8}{
        \begin{tabular}{l|ccccc}
        \toprule
        \textbf{Method}
        & CE & MME & SHOT-IM$++$ & SHOT$++$ & CCT \\
        \midrule
        \textbf{Accuracy} & 57.4 & 62.7 & 65.2 & 66.1 & 68.3 \\
        \bottomrule
        \end{tabular}
    }
    \label{table:ssma}
\end{table}

\vspace{1mm}
\noindent\textbf{Results on SSMA Benchmark.}
In addition to Universal Semi-supervised Model Adaptation~(USMA), we also implement CCT on Semi-supervised Model Adaptation~(SSMA) benchmark. Following~\cite{shotplus}, we conduct experiments on \textit{Office-Home} 1-shot setting using VGG16 as the backbone. The mean accuracy over 12 domain pairs is shown in Table~\ref{table:ssma}. It can be observe that CCT outperforms the current SOTA: SHOT++~\cite{shotplus}, which further demonstrates the versatility of CCT.

\section{Conclusions}
In this paper, we introduce a more realistic and challenging domain adaptation problem called Universal Semi-supervised Model Adaptation (USMA), where i) instead of the source data, only the pre-trained source model is available; ii) the source and target domain do not share the same label set; iii) and there are only a few labeled samples in each class of the target domain.
We show that USMA cannot be resolved by naively fine-tune the pre-trained source model using semi-supervised learning or semi-supervised model adaptation methods. 
Thus, we propose a collaborative consistency training framework that addresses USMA by utilizing the complementary strengths of both the pre-trained source model and its variant pre-trained on target data only. 
Moreover, we propose a two-perspective (\ie, sample-wise and class-wise) consistency regularization that helps to make full advantage of our framework, leading to better performance. Experimental results show that our method surpass existing semi-supervised learning, semi-supervised domain adaptation and semi-supervised model adaptation methods on various benchmark datasets.
% In this paper, we introduce a practical domain adaptation setting called Source-free Universal Semi-supervised Domain Adaptation (SUSDA), where only pre-trained source model is available rather than source data, and the source and target domain do not share the label set. SUSDA requires the model to classify all samples correctly as there are few labeled samples per-category. The SUSDA is challenge since the pre-trained source model is highly source biased, the performance on target private categories is suppressed. To address the issue, we propose to use another model that self-supervised pre-trained on target data only to perform collaboratively consistency training (CCT) with the source model. Specifically, we propose two consistency losses that can effectively make the two models learn from each other, and improve their performance. Experimental results show the efficacy of our proposed CCT under the SUSDA setting.\noindent\textbf{Acknowledgment.} 

\vspace{1mm}
\noindent\textbf{Acknowledgment.} 
The work was supported in part by NSFC with Grant
No. 62293482, the Basic Research Project No. HZQBKCZYZ-2021067 of Hetao ShenzhenHK S\&T Cooperation Zone, the National Key R\&D Program of China with
grant No. 2018YFB1800800, by Shenzhen Outstanding
Talents Training Fund 202002, by Guangdong Research
Projects No. 2017ZT07X152 and No. 2019CX01X104,
by the Guangdong Provincial Key Laboratory of Future
Networks of Intelligence (Grant No. 2022B1212010001),
and by Shenzhen Key Laboratory of Big Data and Artificial Intelligence (Grant No. ZDSYS201707251409055).
It was also partially supported by NSFC62172348, Outstanding Yound Fund of Guangdong Province with No.
2023B1515020055 and Shenzhen General Project with No.
JCYJ20220530143604010.

\newpage
%%%%%%%%% REFERENCES
{\small
\bibliographystyle{ieee_fullname}
\bibliography{egbib}
}

\end{document}